\documentclass[10pt,twocolumn,letterpaper]{article}

\usepackage{cvpr}              

\usepackage{graphicx}
\usepackage{amsmath}
\usepackage{amssymb}
\usepackage{booktabs}

\usepackage[accsupp]{axessibility}

\makeatletter
\@namedef{ver@everyshi.sty}{}
\makeatother
\usepackage{tikz}
\usepackage{pgfplots}
\usepgfplotslibrary{fillbetween}
\usetikzlibrary{shapes,arrows}
\usetikzlibrary{patterns}
\usetikzlibrary{arrows,calc,positioning}

\usetikzlibrary{external}
\pgfplotsset{compat=newest}

\newenvironment{customlegend}[1][]{%
    \begingroup
    \csname pgfplots@init@cleared@structures\endcsname
    \pgfplotsset{#1}%
}{%
    \csname pgfplots@createlegend\endcsname
    \endgroup
}%
\def\addlegendimage{\csname pgfplots@addlegendimage\endcsname}

\usepackage{subcaption}
\usepackage{placeins}
\usepackage{array}

\usepackage{xcolor}

\usepackage[pagebackref,breaklinks,colorlinks]{hyperref}

\usepackage[capitalize]{cleveref}
\crefname{section}{Sec.}{Secs.}
\Crefname{section}{Section}{Sections}
\Crefname{table}{Table}{Tables}
\crefname{table}{Tab.}{Tabs.}

\def\cvprPaperID{*****} 
\def\confName{CVPR}
\def\confYear{2022}

\newcommand{\expect}{\mathbb{E}}

\usepackage{cleveref} 

\begin{document}

\title{A Deeper Look into Aleatoric and Epistemic Uncertainty Disentanglement}

\author{Matias Valdenegro-Toro\\
Department of AI, Bernoulli Institute\\
University of Groningen\\
{\tt\small m.a.valdenegro.toro@rug.nl}
\and
Daniel Saromo Mori\\
Artificial Intelligence Research Group\\
Pontifical Catholic University of Peru\\
{\tt\small daniel.saromo@pucp.pe}
}
\maketitle

\begin{abstract}
    Neural networks are ubiquitous in many tasks, but trusting their predictions is an open issue. Uncertainty quantification is required for many applications, and disentangled aleatoric and epistemic uncertainties are best. In this paper, we generalize methods to produce disentangled uncertainties to work with different uncertainty quantification methods, and evaluate their capability to produce disentangled uncertainties. Our results show that: there is an interaction between learning aleatoric and epistemic uncertainty, which is unexpected and violates assumptions on aleatoric uncertainty, some methods like Flipout produce zero epistemic uncertainty, aleatoric uncertainty is unreliable in the out-of-distribution setting, and Ensembles provide overall the best disentangling quality. We also explore the error produced by the number of samples hyper-parameter in the sampling softmax function, recommending $N > 100$ samples. We expect that our formulation and results help practitioners and researchers choose uncertainty methods and expand the use of disentangled uncertainties, as well as motivate additional research into this topic.
\end{abstract}

\section{Introduction}
\label{sec:intro}

Neural networks are state of the art for many tasks \cite{abiodun2018state}, ranging from Computer Vision \cite{janai2020computer} to Robotics \cite{cebollada2021state}, Natural Language Processing \cite{deng2018deep}, and some Medical applications  \cite{abiodun2018state}. Use cases involving human subjects usually require some safety constraints and, in general, this means a model should produce reasonable estimates of its confidence or uncertainty when making predictions.

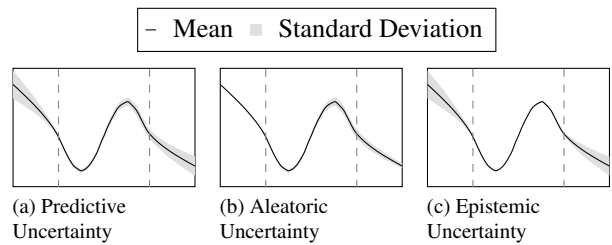
\begin{figure}
    \centering
    \begin{tikzpicture}
        \begin{customlegend}[legend columns = 2,legend style = {column sep=1ex}, legend cell align = left,
            legend entries={Mean, Standard Deviation}]
            \addlegendimage{mark=-,black, only marks}
            \addlegendimage{mark=square*,lightgray!50, only marks}
        \end{customlegend}
    \end{tikzpicture}
    \vspace*{0.5em}
    
    \begin{subfigure}{0.32\linewidth}
        \begin{tikzpicture}
            \begin{axis}[height = 0.14 \textheight, width = 1.5\linewidth, xlabel={}, ylabel={}, xmin = -6.28, xmax = 6.28, grid style=dashed, legend pos = north east, legend style={font=\scriptsize}, tick label style={font=\scriptsize}, ymin =-1.5 , ymax=2.0, xmajorticks=false, ymajorticks=false]
                
                \addplot[name path=upper, draw=none] table[x = x, y expr = \thisrow{pred_mu} + \thisrow{pred_sigma}, col sep = semicolon] {data/toy-regression-15ensembles-disentangled.csv};
                \addplot[name path=lower, draw=none] table[x = x, y expr = \thisrow{pred_mu} - \thisrow{pred_sigma}, col sep = semicolon] {data/toy-regression-15ensembles-disentangled.csv};
                
                \addplot[mark=none, black] table[x = x, y = pred_mu, col sep = semicolon] {data/toy-regression-15ensembles-disentangled.csv};
                
                \addplot [fill=lightgray!50, area legend] fill between[of=upper and lower]; 
                
                \draw[dashed, gray] (3.1415, -2) -- (3.1415, 2);
                \draw[dashed, gray] (-3.1415, -2) -- (-3.1415, 2);
            \end{axis}		
        \end{tikzpicture}
        \caption{Predictive\\Uncertainty}
    \end{subfigure}
    \begin{subfigure}{0.32\linewidth}
        \begin{tikzpicture}
            \begin{axis}[height = 0.14 \textheight, width = 1.5\linewidth, xlabel={}, ylabel={}, xmin = -6.28, xmax = 6.28, grid style=dashed, legend pos = north east, legend style={font=\scriptsize}, tick label style={font=\scriptsize}, ymin =-1.5 , ymax=2.0, xmajorticks=false, ymajorticks=false]
                
                \addplot[name path=upper, draw=none, forget plot] table[x = x, y expr = \thisrow{pred_mu} + \thisrow{pred_sigma_ale}, col sep = semicolon] {data/toy-regression-15ensembles-disentangled.csv};
                \addplot[name path=lower, draw=none, forget plot] table[x = x, y expr = \thisrow{pred_mu} - \thisrow{pred_sigma_ale}, col sep = semicolon] {data/toy-regression-15ensembles-disentangled.csv};
                
                \addplot[mark=none, black] table[x = x, y = pred_mu, col sep = semicolon] {data/toy-regression-15ensembles-disentangled.csv};
                                                              
                \addplot [fill=lightgray!50, area legend] fill between[of=upper and lower]; 
                
                \draw[dashed, gray] (3.1415, -2) -- (3.1415, 2);
                \draw[dashed, gray] (-3.1415, -2) -- (-3.1415, 2);       
            \end{axis}		
        \end{tikzpicture}
        \caption{Aleatoric\\Uncertainty}
    \end{subfigure} 
    \begin{subfigure}{0.32\linewidth}
        \begin{tikzpicture}
            \begin{axis}[height = 0.14 \textheight, width = 1.5\linewidth, xlabel={}, ylabel={}, xmin = -6.28, xmax = 6.28, grid style=dashed, legend pos = north east, legend style={font=\scriptsize}, tick label style={font=\scriptsize}, ymin =-1.5 , ymax=2.0, xmajorticks=false, ymajorticks=false]
                
                \addplot[name path=upper, draw=none] table[x = x, y expr = \thisrow{pred_mu} + \thisrow{pred_sigma_epi}, col sep = semicolon] {data/toy-regression-15ensembles-disentangled.csv};
                \addplot[name path=lower, draw=none] table[x = x, y expr = \thisrow{pred_mu} - \thisrow{pred_sigma_epi}, col sep = semicolon] {data/toy-regression-15ensembles-disentangled.csv};
                
                \addplot[mark=none, black] table[x = x, y = pred_mu, col sep = semicolon] {data/toy-regression-15ensembles-disentangled.csv};
                               
                \addplot[fill=lightgray!50] fill between[of=upper and lower]; 
                                
                \draw[dashed, gray] (3.1415, -2) -- (3.1415, 2);
                \draw[dashed, gray] (-3.1415, -2) -- (-3.1415, 2);
            \end{axis}		
        \end{tikzpicture}
        \caption{Epistemic\\Uncertainty}
    \end{subfigure}
    \caption{Example of uncertainty disentanglement in toy regression of a sinusoid, produced using an ensemble of 15 neural networks, with standard deviation being computed across the ensemble predictions. Predictive uncertainty is decomposed into aleatoric and epistemic uncertainty, where aleatoric is Gaussian noise added to the data, and epistemic is higher in out of distribution inputs (indicated by the dashed bars for $x < -\pi$ and $x > \pi$).}
    \label{fig:ensemble_example}
\end{figure}

There are two kinds of uncertainty \cite{uncertaintyGal2016, hullermeier2021aleatoric, kendall2017uncertainties}: aleatoric or data uncertainty, and epistemic or model uncertainty. These uncertainties are usually combined and predicted as a single value, called predictive uncertainty \cite{uncertaintyGal2016}. Recovering the two components of uncertainty is helpful for certain applications. For example, in active learning, epistemic uncertainty can guide the selection of samples to label, but aleatoric uncertainty should be ignored. In robot perception (segmentation, object detection), it is helpful to separate data uncertainty from model uncertainty, for purposes of out-of-distribution detection, and to ignore outliers and noise. Figure \ref{fig:ensemble_example} shows an example of disentangled uncertainties for a toy regression example.

There are methods to disentangle aleatoric and epistemic uncertainty for some machine learning models \cite{depeweg2018decomposition, senge2014reliable, depewegModelingEpistemic}. For deep neural networks, Kendall and Gal. \cite{kendall2017uncertainties} define a general disentangling model, but it is mostly defined for a base model using MC-Dropout \cite{gal2016dropout}. In this paper, we generalize this formulation to allow disentanglement across multiple methods of uncertainty estimation (Like Ensembles, Flipout, etc).

We make an experimental comparison between different uncertainty quantification methods relative to their capacity for disentangling aleatoric and epistemic uncertainty. We tested those techniques in regression and classification tasks (on the FER+ dataset), and explore the interaction between both sources of uncertainty.

Overall, we find that for the purpose of disentangling, aleatoric and epistemic uncertainty do interact, which is unexpected, as only epistemic uncertainty should interact with the model, and not aleatoric uncertainty. In particular with Flipout, outputting only aleatoric uncertainty and zero epistemic uncertainty, even for out of distribution cases, which we believe is an anomaly. We also find that aleatoric uncertainty estimation is unreliable in out-of-distribution settings, particularly for regression, with constant aleatoric variances being output by a model. Our results show that Ensembles have the best uncertainty and disentangling behavior for both classification and regression, and the $\beta$-NLL loss \cite{seitzer2022on} improves both aleatoric and epistemic uncertainty quantification, while Seitzer et al. had explored only its use for aleatoric uncertainty.

The contributions of this paper are a generalization of methods to disentangle aleatoric and epistemic uncertainty produced by a machine learning model across different uncertainty quantification methods, which were originally proposed by Kendall and Gal. \cite{kendall2017uncertainties} but only for MC-Dropout; and a comparison between dropout, dropconnect, ensembles, and flipout, about their disentangling quality across a regression and classification tasks. We also explore setting the number of samples in the sampling softmax function, recommending the use $N = 100$ samples to prevent incorrect classification due to approximation error.

\section{Related Work}


As we mentioned before, separating the total uncertainty into its epistemic and aleatoric components is necessary for specific applications. There are some approaches for achieving this process \cite{hullermeier2021aleatoric}. Depeweg et al. \cite{depeweg2018decomposition} presented a method for measuring the total and aleatoric uncertainties, and then they calculated the epistemic element by subtracting those values. Nevertheless, they tested their proposal only in regression and reinforcement learning tasks. Alternatively, Kendall and Gal \cite{kendall2017uncertainties} developed another approach for independently calculating both uncertainty components using MC-Dropout. The authors tested their method at regression and classification problems. In this work we expand this disentanglement method to consider other uncertainty quantification methods.


On the other hand, we cannot know the exact probability distribution of the inputs that will be given to the trained model. Hence, quantify its uncertainty, we pass it samples from the training dataset, \ie following the Empirical Risk Minimization (ERM) principle \cite{vapnik1991principles}. The most common sampling methods are: Monte Carlo Dropout \cite{gal2016dropout, uncertaintyGal2016} (which turns off some activations at each sample passing to estimate the prediction uncertainty), Monte Carlo DropConnect \cite{mobiny2021dropconnect} (which turns off weights instead of activations, and whose authors reported it to be capable of achieving better results than MC Dropout \cite{mobiny2021dropconnect}), estimation with deep ensembles \cite{lakshminarayanan2017simple, ovadia2019can} (which blends the predictions from networks with different weight initializations taken from the same probability distribution), Flipout-based variational inference \cite{wen2018flipout} (which samples weight perturbations inside a mini-batch), and Markov Chain Monte Carlo \cite{kupinski2003ideal} (which uses sequential drawings from a stochastic distribution to estimate the exact posterior); a review of these techniques can be found in \cite{uqreview2021}. 



\section{Uncertainty Disentanglement}
Many uncertainty quantification methods can be categorized into sampling-based and ensemble-based, where ensembling can be seen conceptually as a way of sampling. In this section we generalize the method proposed by Kendall and Gall \cite{kendall2017uncertainties}.
\subsection{Regression}
Assume we have a model with uncertainty that outputs two quantities: the mean $\mu_i(\mathbf{x})$ and variance $\sigma^2_i(\mathbf{x})$, where $i \in [1, M]$ is an index for different samples or ensembles. For the purpose of uncertainty quantification, we usually sample weights from a weight distribution $\theta \sim p(\theta | \mathbf{x}, y)$, which produce different predictions for $\mu$ and $\sigma^2$, that correspond to samples of the (approximate) predictive posterior distribution of the model.

\newcommand{\plotbinaryprobsedge}[2]{
    \begin{tikzpicture}
        \begin{axis}[height = 0.15 \textheight, width = 0.35 \linewidth, grid style=dashed, tick label style={font=\scriptsize}, ymin =0.0 , ymax=1.0, ytick={0.2, 0.4, 0.6, 0.8, 1.0}, xticklabels={,,}, xmajorticks=false, ylabel={Probability}]
            
            \addplot[mark=none, gray, ybar, fill] coordinates {(0, #1) (1, #2)};
        \end{axis}		
    \end{tikzpicture}
}

\newcommand{\plotbinaryprobsbottom}[2]{
    \begin{tikzpicture}
        \begin{axis}[height = 0.15 \textheight, width = 0.35 \linewidth, grid style=dashed, tick label style={font=\scriptsize}, ymin =0.0 , ymax=1.0, xtick=data, yticklabels={,,}, ymajorticks=false, xlabel={Class}]
            
            \addplot[mark=none, gray, ybar, fill] coordinates {(0, #1) (1, #2)};
        \end{axis}		
    \end{tikzpicture}
}

\newcommand{\plotbinaryprobsedgebottom}[2]{
    \begin{tikzpicture}
        \begin{axis}[height = 0.15 \textheight, width = 0.35 \linewidth, grid style=dashed, tick label style={font=\scriptsize}, ymin =0.0 , ymax=1.0, xtick=data, ytick={0.2, 0.4, 0.6, 0.8, 1.0}, xlabel={Class}, ylabel={Probability}]
            
            \addplot[mark=none, gray, ybar, fill] coordinates {(0, #1) (1, #2)};
        \end{axis}		
    \end{tikzpicture}
}

\newcommand{\plotbinaryprobs}[2]{
    \begin{tikzpicture}
        \begin{axis}[height = 0.15 \textheight, width = 0.35 \linewidth, grid style=dashed, tick label style={font=\scriptsize}, ymin =0.0 , ymax=1.0, xtick=data, yticklabels={,,}, xticklabels={,,}, xmajorticks=false, ymajorticks=false]
            
            \addplot[mark=none, gray, ybar, fill] coordinates {(0, #1) (1, #2)};
        \end{axis}		
    \end{tikzpicture}
}

\begin{figure}[t]
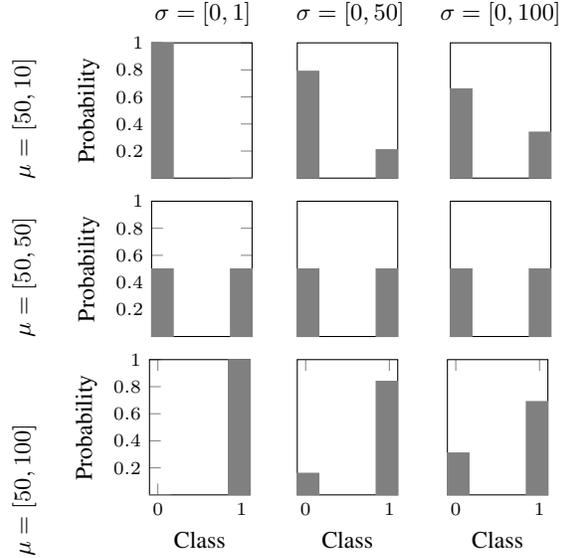

    \small
    \begin{tabular}{lccc}
        &\hspace*{3em} $\sigma = [0, 1]$		& $\sigma = [0, 50]$		& $\sigma = [0, 100]$\\
        \rotatebox{90}{$\mu = [50, 10]$} & \plotbinaryprobsedge{1.0}{0.0} 	& \plotbinaryprobs{0.79}{0.21}	& \plotbinaryprobs{0.66}{0.34}\\
        \rotatebox{90}{$\mu = [50, 50]$} & \plotbinaryprobsedge{0.5}{0.5} 	& \plotbinaryprobs{0.5}{0.5}	& \plotbinaryprobs{0.5}{0.5}\\
        \rotatebox{90}{$\mu = [50, 100]$} & \plotbinaryprobsedgebottom{0.0}{1.0} 	& \plotbinaryprobsbottom{0.16}{0.84}	& \plotbinaryprobsbottom{0.31}{0.69}
    \end{tabular}    
    \caption{Visualizations of binary probability distributions produced by sampling softmax with different logit Gaussian distributions $\mathcal{N}(\mu, \sigma^2)$ with given mean $\mu$ and standard deviation $\sigma$.}
    \label{fig:sample_softmax_examples}
\end{figure}

These samples are usually combined into a single Gaussian mixture distribution $p(y \, | \, \mathbf{x})$ using:
\begin{align}
    p(y \, | \, \mathbf{x}) &\sim \mathcal{N}(\mu_*(\mathbf{x}), \sigma^2_*(\mathbf{x}))\\
    \mu_*(\mathbf{x}) &= M^{-1} \sum_i \mu_i(\mathbf{x})\\
    \sigma^2_*(\mathbf{x}) &= M^{-1} \sum_i (\sigma^2_i(\mathbf{x}) + \mu^2_i(\mathbf{x})) - \mu^2_*(\mathbf{x})
 \end{align}

\begin{figure}[!t]
    \centering
    \begin{tikzpicture}[style={align=center, minimum height=0.5cm, minimum width = 2.6cm}]
        \node[] (dummy) {};
        \node[rounded corners, draw, above=1.5em of dummy] (sampSoftmax) {{Sampling Softmax}};
        \node[above=1.0em of sampSoftmax] (probs) {{Probabilities}};
        
        \node[rounded corners, draw, left=0.001em of dummy] (objClassifier) {{Logit Mean}};
        
        \node[rounded corners, draw, right=0.001em of dummy] (objDetector) {{Logit Variance}};
        
        \node[below=1.5em of dummy] (fc) {{Feature Vector}};
        
        \draw[-latex] (fc) -- (objDetector);
        \draw[-latex] (fc) -- (objClassifier);
        \draw[-latex] (objClassifier) -- (sampSoftmax);
        \draw[-latex] (objDetector) -- (sampSoftmax);
        \draw[-latex] (sampSoftmax) -- (probs);
    \end{tikzpicture}
    \caption{Computational graph for the sampling softmax function, showing the two fully connected layers that produce logit mean and variance, and how they relate to the final probabilities through the sampling softmax function.}
    \label{fig:sampling_softmax_graph}
\end{figure}
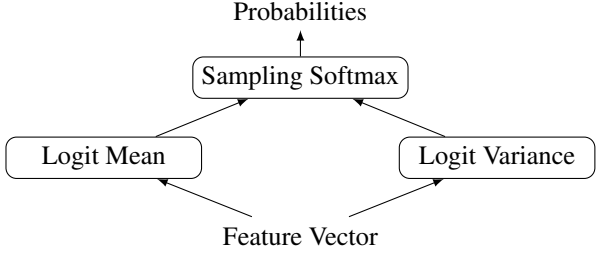

For the predictive variance $\sigma^2_*(x)$, this can be decomposed into aleatoric and epistemic uncertainty, by rewriting as:

\begin{table}[h!]
\addtolength{\tabcolsep}{-4pt}
\begin{tabular}{ccccc}
$\sigma^2_*(\mathbf{x})$ & $=$ & $\displaystyle M^{-1} \sum_i \sigma^2_i(\mathbf{x})$ & $+$ & $\displaystyle M^{-1} \sum_i \mu^2_i(\mathbf{x})- \mu^2_*(\mathbf{x})$\\[1ex]
& $=$ &  $\expect_i [\sigma_i^2(\mathbf{x})]$ & $+$ & $\expect_i [\mu_i^2(\mathbf{x})] - \expect_i [\mu_i(\mathbf{x})]^2$\\[1ex]
& $=$ & $\underbrace{\expect_i [\sigma_i^2(\mathbf{x})]}_{\text{Aleatoric Uncertainty}}$ & $+$ & $\underbrace{\text{Var}_i[\mu_i(\mathbf{x})]}_{\text{Epistemic Uncertainty}}$
\end{tabular}
\end{table}

This derivation indicates that across forward pass samples, the mean of the variances represents aleatoric uncertainty, while the variance of the means corresponds to epistemic uncertainty. Also, this formulation can be derived by using the law of total variance.

The variance heads $\sigma^2(x)$ of a model can be trained using the Gaussian negative log-likelihood loss for a sample indexed by $n$ with input $\mathbf{x}_n$ and label $y_n$:
\begin{equation}
    L_{NLL}(y_n, \mathbf{x}_n) = \frac{\log \sigma^2_i(\mathbf{x}_n)}{2} + \frac{(\mu_i(\mathbf{x}_n) - y_n)^2}{2 \sigma^2_i(\mathbf{x}_n)} .
    \label{eq:gaussian_nll}
\end{equation}

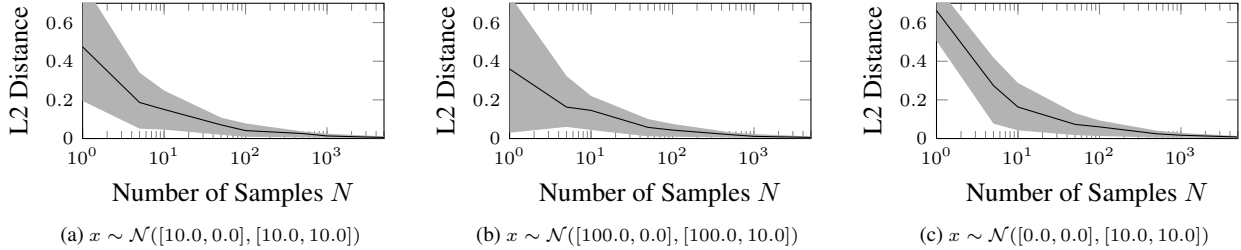
\begin{figure*}
    \centering
    \begin{subfigure}{0.32\textwidth}
        \begin{tikzpicture}
            \begin{axis}[height = 0.15 \textheight, width = \linewidth, xlabel={Number of Samples $N$}, ylabel={L2 Distance}, xmin = 1, xmax = 5000, grid style=dashed, legend pos = north east, legend style={font=\scriptsize}, tick label style={font=\scriptsize}, ymin =0.0 , ymax=0.7, xmode=log]
                
                \addplot+[mark = none, black] table[x  = num_samples, y  = mean_error, col sep = semicolon] {data/sampling-softmax-error-means10.-0.-stds10.-10.csv};
                
                \addplot[name path=upper, draw=none] table[x = num_samples, y expr = \thisrow{mean_error} + \thisrow{std_error}, col sep = semicolon] {data/sampling-softmax-error-means10.-0.-stds10.-10.csv};
                \addplot[name path=lower, draw=none] table[x = num_samples, y expr = \thisrow{mean_error} - \thisrow{std_error}, col sep = semicolon] {data/sampling-softmax-error-means10.-0.-stds10.-10.csv};
                
                \addplot [fill=darkgray!40] fill between[of=upper and lower]; 
            \end{axis}        
        \end{tikzpicture}
        \caption{\scriptsize $x \sim \mathcal{N}([10.0, 0.0], [10.0, 10.0])$}
    \end{subfigure}
    \begin{subfigure}{0.32\textwidth}
        \begin{tikzpicture}
            \begin{axis}[height = 0.15 \textheight, width = \linewidth, xlabel={Number of Samples $N$}, ylabel={L2 Distance}, xmin = 1, xmax = 5000, grid style=dashed, legend pos = north east, legend style={font=\scriptsize}, tick label style={font=\scriptsize}, ymin =0.0 , ymax=0.7, xmode=log]
                
                \addplot+[mark = none, black] table[x  = num_samples, y  = mean_error, col sep = semicolon] {data/sampling-softmax-error-means100.-0.-stds100.-10.csv};
                
                \addplot[name path=upper, draw=none] table[x = num_samples, y expr = \thisrow{mean_error} + \thisrow{std_error}, col sep = semicolon] {data/sampling-softmax-error-means100.-0.-stds100.-10.csv};
                \addplot[name path=lower, draw=none] table[x = num_samples, y expr = \thisrow{mean_error} - \thisrow{std_error}, col sep = semicolon] {data/sampling-softmax-error-means100.-0.-stds100.-10.csv};
                
                \addplot [fill=darkgray!40] fill between[of=upper and lower]; 
            \end{axis}
        \end{tikzpicture}
        \caption{\scriptsize $x \sim \mathcal{N}([100.0, 0.0], [100.0, 10.0])$}
    \end{subfigure}
    \begin{subfigure}{0.32\textwidth}
        \begin{tikzpicture}
            \begin{axis}[height = 0.15 \textheight, width = \linewidth, xlabel={Number of Samples $N$}, ylabel={L2 Distance}, xmin = 1, xmax = 5000, grid style=dashed, legend pos = north east, legend style={font=\scriptsize}, tick label style={font=\scriptsize}, ymin =0.0 , ymax=0.7, xmode=log]
                
                \addplot+[mark = none, black] table[x  = num_samples, y  = mean_error, col sep = semicolon] {data/sampling-softmax-error-means0.-0.-stds10.-10.csv};
                
                \addplot[name path=upper, draw=none] table[x = num_samples, y expr = \thisrow{mean_error} + \thisrow{std_error}, col sep = semicolon] {data/sampling-softmax-error-means0.-0.-stds10.-10.csv};
                \addplot[name path=lower, draw=none] table[x = num_samples, y expr = \thisrow{mean_error} - \thisrow{std_error}, col sep = semicolon] {data/sampling-softmax-error-means0.-0.-stds10.-10.csv};
                
                \addplot [fill=darkgray!40] fill between[of=upper and lower]; 
            \end{axis}		
        \end{tikzpicture}
        \caption{\scriptsize $x \sim \mathcal{N}([0.0, 0.0], [10.0, 10.0])$}
    \end{subfigure}        
    \caption{Error produced by different number of samples through the sampling softmax function, measured as L2 distance between probabilities with given number of samples and the best approximation, with three different Gaussian logit distributions. Shaded areas represent one-$\sigma$ error bars.}
    \label{fig:error_sampling_softmax} 
\end{figure*}

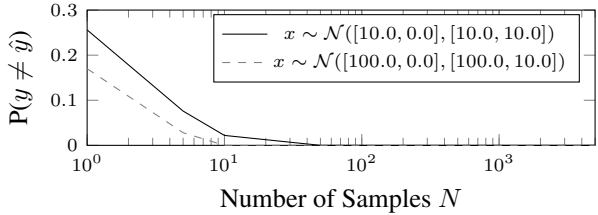
\begin{figure}
    \begin{tikzpicture}
        \begin{axis}[height = 0.15 \textheight, width = \linewidth, xlabel={Number of Samples $N$}, ylabel={P($y \neq \hat{y}$)}, xmin = 1, xmax = 5000, grid style=dashed, legend pos = north east, legend style={font=\scriptsize}, tick label style={font=\scriptsize}, ymin =0.0 , ymax=0.3, xmode=log]
            
            \addplot+[mark = none, black] table[x  = num_samples, y  =  mean_miss, col sep = semicolon] {data/sampling-softmax-error-means10.-0.-stds10.-10.csv};
            \addlegendentry{$x \sim \mathcal{N}([10.0, 0.0], [10.0, 10.0])$}
            \addplot+[mark = none, gray, dashed] table[x  = num_samples, y  =  mean_miss, col sep = semicolon] {data/sampling-softmax-error-means100.-0.-stds100.-10.csv};
            \addlegendentry{$x \sim \mathcal{N}([100.0, 0.0], [100.0, 10.0])$}
        \end{axis}        
    \end{tikzpicture}
    \caption{Probability of misclassification produced by different number of samples through the sampling softmax function and the best approximation, with two different Gaussian logit distributions.}
    \label{fig:misclass_sampling_softmax}
\end{figure}

This loss is also called variance attenuation. Nevertheless, this loss is known to have issues underestimating the variance head. Hence, an alternative called $\beta$-NLL \cite{seitzer2022on} has been proposed to minimize these issues:
\begin{equation}
    L_{\beta-NLL}(y_n, \mathbf{x}_n) = \text{stop}(\sigma^{2\beta}) \,\, L_{NLL}(y_n, \mathbf{x}_n) .
    \label{eq:gaussian_betanll}
\end{equation}

Where $\text{stop}()$ is the stop gradient operation, that prevents gradients from flowing through the operation inside the parenthesis. This loss makes the predicted variance act as a weight for each data point, putting more weight into larger variances. The parameter $\beta$ controls the strength of this weighting.
\subsection{Classification}
\label{refSectionClasification}

In classification problems, it is slightly more difficult to separately model aleatoric and epistemic uncertainty. Kendall and Gal \cite{kendall2017uncertainties} proposed to make a custom Softmax activation layer that models logits $z$ with uncertainty (Gaussian mean and variance), and uses sampling (with $N$ samples) to pass the Gaussian logit distribution $\hat{\mathbf{z}}$ through the softmax activation to produce $p(y | \mathbf{x})$. We call this function the \textit{sampling softmax function} (Eq \ref{eq:ssoftmax1} and \ref{eq:ssoftmax2}).
\begin{align}
    \hat{\mathbf{z}}_j &\sim \mathcal{N}(\mu(\mathbf{x}), \sigma^2(\mathbf{x}))\label{eq:ssoftmax1} \\
    p(y | \mathbf{x}) &= N^{-1} \sum_j \text{softmax}(\hat{\mathbf{z}}_j) \qquad j \in [1, N] \label{eq:ssoftmax2}
\end{align}
Then at inference time, we again assume that the uncertainty method uses sampling through forwarding passes or ensembling a model on the $i$ axis (with $i \in [1, M]$). Then, aleatoric $\sigma^2_{\text{Ale}}$ and epistemic $\sigma^2_{\text{Epi}}$ uncertainty logits can be computed
\begin{align}
    \sigma^2_{\text{Ale}}(\mathbf{x}) &= \expect_i [\sigma_i^2(\mathbf{x})] \qquad \sigma^2_{\text{Epi}}(\mathbf{x}) &= \text{Var}_i [\mu_i(\mathbf{x})].
\end{align}
Note that these are logits, not probabilities. Passing each corresponding logit through a softmax function can produce probabilities, from where entropy is a possible metric to obtain a scalar uncertainty measure:
\begin{align*}
    p_{\text{Ale}}(y | \mathbf{x}) &= \text{sampling\_softmax}(\mu(\mathbf{x}), \sigma^2_{\text{Ale}}(\mathbf{x}))\\
    H_{\text{Ale}}(y | \mathbf{x}) &= \text{entropy}(p_{Ale}(y | \mathbf{x}))\\    
    p_{\text{Epi}}(y | \mathbf{x}) &= \text{sampling\_softmax}(\mu(\mathbf{x}), \sigma^2_{\text{Epi}}(\mathbf{x}))\\
    H_{\text{Epi}}(y | \mathbf{x}) &= \text{entropy}(p_{Epi}(y | \mathbf{x}))
\end{align*}
Where $\mu(x) = M^{-1} \sum_i \mu_i(x)$ is the predictive mean and entropy is the standard Shannon entropy defined as $\text{entropy}(p) = -\sum_i p_i \log p_i$.

It should be noted that unlike the case of regression, in classification, disentangling uncertainties and transforming them into probabilities does not mean that epistemic and aleatoric probabilities or entropy will sum to the predictive probabilities or entropy. Only the logits can be summed to obtain predictive logits.

Figure \ref{fig:sample_softmax_examples} shows the behavior of the sampling softmax function with different logit distributions. The behavior is not so intuitive, particularly when the means of both logit distributions are the same. When the means are different, the logit variances play a more significant role and affect the final probabilities. Figure \ref{fig:sampling_softmax_graph} shows the computational graph of this function.

\begin{figure*}
    \includegraphics[width=\textwidth]{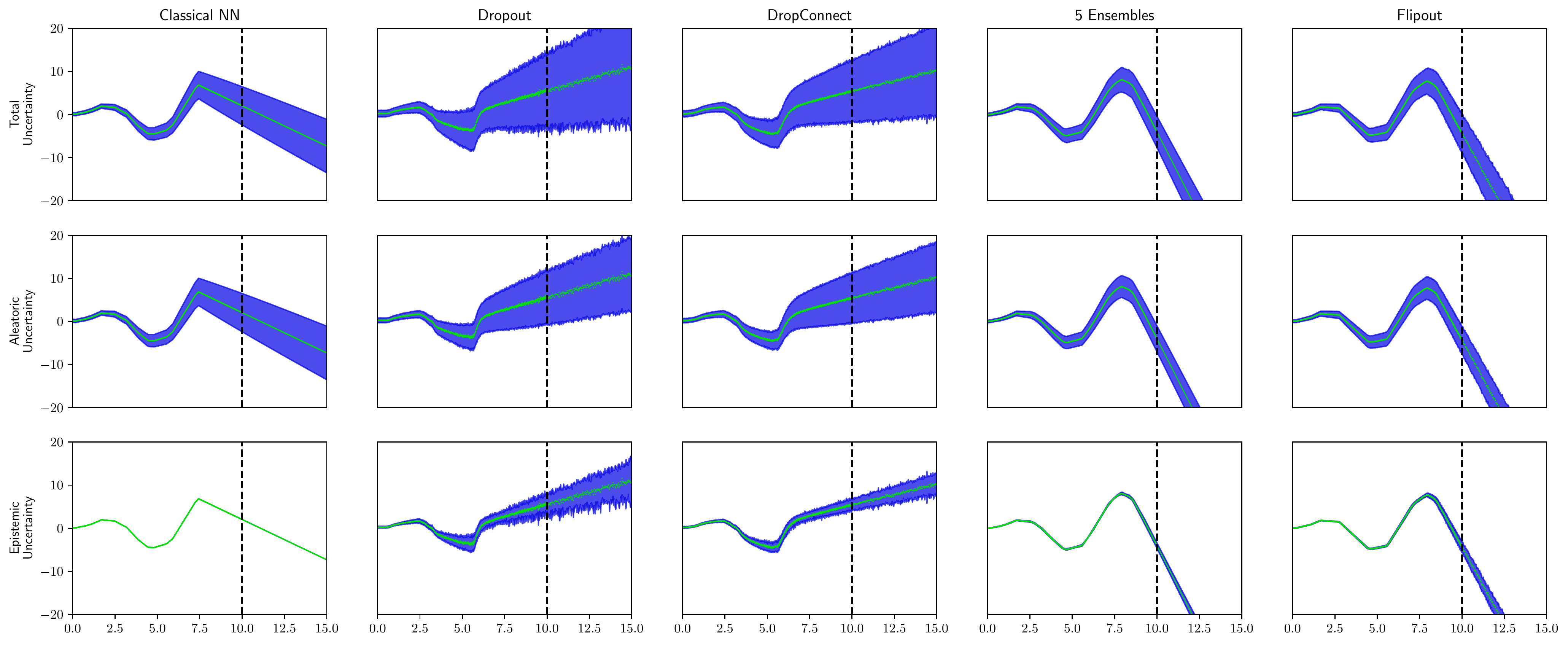}
    \caption{Comparison of four methods for disentangling uncertainty in a toy sinusoid problem with aleatoric uncertainty (homoscedastic and heteroscedatic), using the variance atenuation (NLL) loss. Dropout and DropConnect overestimate epistemic uncertainty on the training set, while ensembles and flipout have low uncertainty outside the training set.}
    \label{fig:toy_regression_nll}
\end{figure*}

\begin{figure*}[!tb]
    \includegraphics[width=\textwidth]{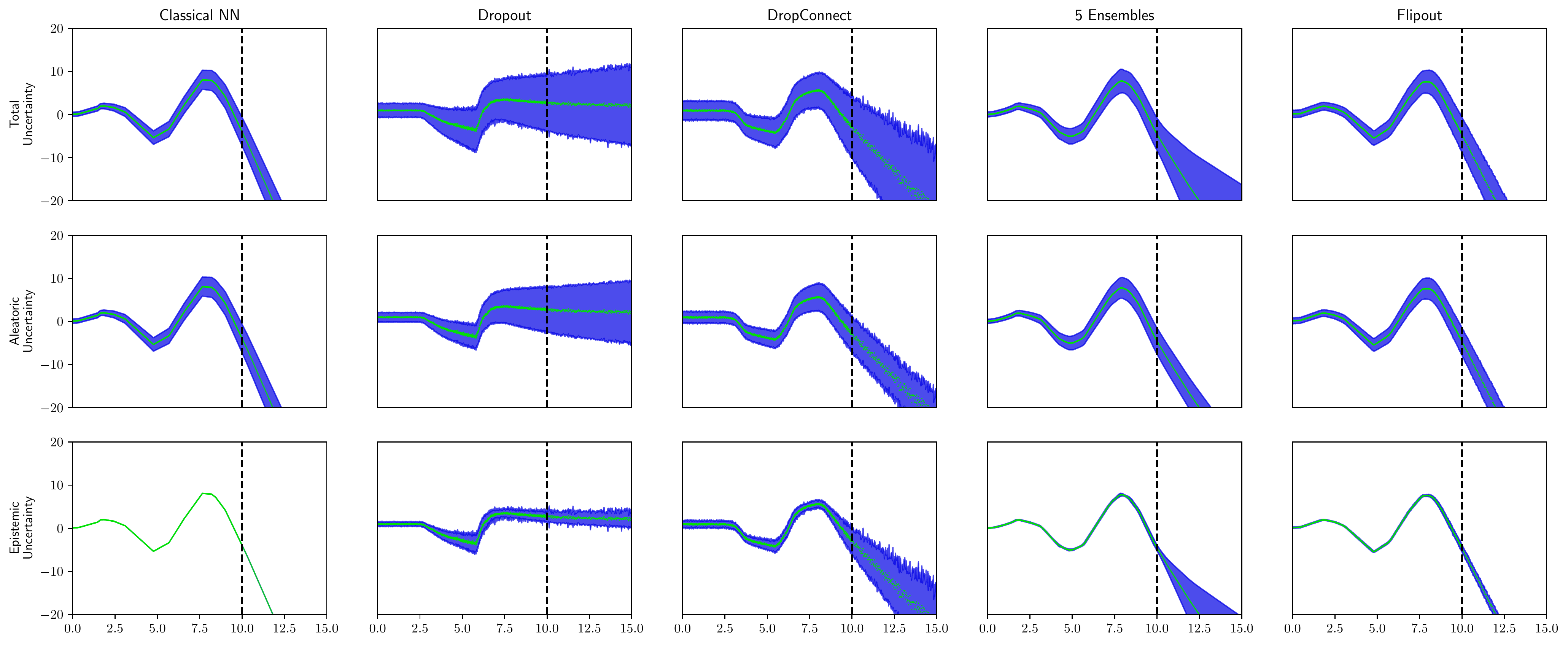}
    \caption{Comparison of four uncertainty methods in disentangling uncertainty for a toy sinusoid problem with aleatoric uncertainty (homoscedastic and heteroscedatic), using the $\beta$-NLL loss. Dropout and DropConnect overestimate epistemic uncertainty on the training set, while ensembles has increased epistemic uncertainty outside the training set.}
    \label{fig:toy_regression_betanll}
\end{figure*}

\subsection{Tuning the Number of Samples}

Kendall and Gal \cite{kendall2017uncertainties} do not provide information on how the number of samples $N$ for the sampling softmax function should be selected. This parameter controls a trade-off between computational performance and error in the estimated probabilities. We evaluate this parameter by estimating the L2 error between $N = 100000$ and a variable value of $N$. These results are shown in Figure \ref{fig:error_sampling_softmax}. We also compute the chance of producing a misclassification due to sampling error in Figure \ref{fig:misclass_sampling_softmax}. From these results, at least $N = 100$ samples are required to obtain zero classification error.

\section{Experimental Comparison}

In this section, we compare different uncertainty estimation methods in terms of their ability to disentangle aleatoric from epistemic uncertainty.

\subsection{Uncertainty Methods}

\textbf{MC-Dropout}. Dropout sets random activations in a layer to zero, and it is intended as a regularizer that is only applied during training. MC-Dropout \cite{gal2016dropout} enables the activation drop during test/inference time, and the model becomes stochastic, where each forward pass produces one sample from the Bayesian posterior distribution \cite{gal2016dropout}. We use dropout layers with a drop probability $p = 0.25$.

\textbf{MC-DropConnect}. DropConnect is conceptually similar to Dropout, randomly dropping weights to zero instead of activations, with a similar regularization effect. MC-DropConnect enables dropping weights at inference time, which also produces samples from the Bayesian posterior distribution \cite{mobiny2021dropconnect}. We use DropConnect layers with drop probability $p = 0.10$. 

\textbf{Ensembles}. Ensembles consist of training multiple copies of the same architecture (with different instances of a random weight initialization) and then combining their outputs, which usually produces a better model. Lakshminarayanan \cite{lakshminarayanan2017simple} demonstrated that ensembles also have good uncertainty quantification properties. We use an ensemble of $M = 5$ neural networks.

\textbf{Flipout}. Flipout-based variational inference is a popular method which models weights as an approximate Gaussian distribution \cite{blundell2015weight}, where the kernel and bias matrices are Gaussian distributed. This process generates a stochastic model. Flipout \cite{wen2018flipout} is used to reduce the training process variance, improving learning stability and performance. We use Flipout in multiple layers with a disabled prior, and biases are scalars instead of distributions.

For evaluation we take $M = 20$ forward passes of each method. After that, we compute the mean of probabilities for classification, and the mean and standard deviation for regression, both across forward pass samples. The sampling softmax layer uses $N = 100$ samples for the classification task. As baseline, we also train a neural network without epistemic uncertainty quantification, which we denote as Classical NN. For the regression setting this network uses a mean and variance output heads, to be able to estimate aleatoric uncertainty.

\subsection{Toy Regression}

We first evaluate a simple regression problem, generating a dataset by sampling the following function:

\begin{equation}
    f(x) = x \sin(x) + \epsilon_1 x + \epsilon_2
\end{equation}

Where $\epsilon_1, \epsilon_2 \sim \mathcal{N}(0, 0.3)$. This function has both homoscedatic ($\epsilon_2$) and heteroscedatic ($\epsilon_1$) aleatoric uncertainty, and a model has to estimate both. We produce 1000 samples for $x \in [0, 10]$ as a training set, and an out-of-distribution dataset is built with 200 samples for $x \in [10, 15]$.

We train models using the Gaussian NLL (Eq \ref{eq:gaussian_nll}) and $\beta$-Gaussian NLL (Eq \ref{eq:gaussian_betanll}) with $\beta = 0.5$. These results are available in Figures \ref{fig:toy_regression_nll} and \ref{fig:toy_regression_betanll}. The training data that was used to produce these results is available in the Appendix, Figure \ref{fig:toy_regression_data}.

Our results show that aleatoric uncertainty estimation is unreliable in the out-of-distribution setting (values $x > 10$ in our experiments) for both losses. Ensembles and Flipout in particular produce constant aleatoric uncertainty in the out-of-distribution areas, indicating that they did not learn the general noise pattern in the data (where the noise variance $\epsilon_1$ would continue increasing), and it seems these methods only learned the term $\epsilon_2$ correctly. In contrast, it is interesting to see that the Classical NN did learn that aleatoric uncertainty is increasing even in the out-of-distribution areas.

For the Classical NN, Ensemble, and Flipout, the $\beta$-NLL loss allows us to obtain a similar variance as with the NLL loss, at both aleatoric and epistemic uncertainties. However, both losses generate similar epistemic uncertainty with the dropout method. Nevertheless, for Flipout, epistemic uncertainty is reduced when using $\beta$-NLL (in comparison with the standard NLL). Ensembles, in contrast, have increasing epistemic uncertainty in the output distribution setting with $\beta$-NLL, while with NLL epistemic uncertainty is smaller.

Noticeably, the best methods for epistemic uncertainty are Flipout and Ensembles, with different combinations of losses; but overall, for disentanglement, DropConnect seems to perform very well with $\beta$-NLL.


Altogether, these results show that disentanglement quality relates to both the uncertainty quantification model and the NLL loss choice.

\begin{table}
    \centering
    \begin{tabular}{lll}
        \toprule
        UQ Method		& Test Accuracy (\%)	& Loss\\
        \midrule
        Baseline		& 74.8 \%				& 0.218\\
        Dropout			& 76.7 \%				& 0.124\\
        DropConnect		& 74.3 \%				& 0.167\\
        Flipout			& 70.1 \%				& 0.150\\
        Ensemble		& 77.8 \%				& 0.130\\
        \bottomrule
    \end{tabular}
    \caption{Accuracy and loss metrics for FER+ facial emotion classification with uncertainty quantification.}
    \label{fig:ferplus_metrics}
\end{table}

\newcommand{\plotprobs}[9][0.0]{
    \begin{tikzpicture}
        \begin{axis}[height = 0.12 \textheight, width = 0.20 \textwidth, xlabel={}, ylabel={}, xmin = 1, xmax = 8.5, grid style=dashed, legend pos = north east, legend style={font=\scriptsize}, tick label style={font=\scriptsize}, ymin =0.0 , ymax=1.0, yticklabels={,,}, xticklabels={,,}, xmajorticks=false, ymajorticks=false]
            
            \addplot+[mark=none, red, ybar, fill] coordinates { (1, #2) };
            \addplot+[mark=none, magenta, ybar, fill] coordinates { (2, #3) };
            \addplot+[mark=none, orange, ybar, fill] coordinates { (3, #4) };
            \addplot+[mark=none, violet, ybar, fill] coordinates { (4, #5) };
            \addplot+[mark=none, green, ybar, fill] coordinates { (5, #6) };
            \addplot+[mark=none, lime, ybar, fill] coordinates { (6, #7) };
            \addplot+[mark=none, teal, ybar, fill] coordinates { (7, #8) };
            \addplot+[mark=none, blue, ybar, fill] coordinates { (8, #9) };
            
            \node[] at (axis cs: 7, 0.8) {#1};
        \end{axis}		
    \end{tikzpicture}
}

\newcommand{\inpimage}{\rotatebox{90}{\parbox{2em}{Input\\Image}}}
\newcommand{\trueprobs}{\rotatebox{90}{\parbox{2em}{True\\Probs}}}
\newcommand{\predprobs}{\rotatebox{90}{\parbox{2em}{Predicted\\Probs}}}
\newcommand{\aleprobs}{\rotatebox{90}{\parbox{2em}{Aleatoric\\Probs}}}
\newcommand{\epiprobs}{\rotatebox{90}{\parbox{2em}{Epistemic\\Probs}}}

\begin{figure*}[!h]
    \footnotesize
    \centering
    \begin{tabular}{llllllll}
        & &      				& Baseline & Dropout & DropConnect & Flipout & Ensembles\\
       \inpimage & \includegraphics[width=0.06\linewidth]{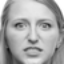}
        & \predprobs			& \plotprobs[0.311]{0.93}{0.05}{0.01}{0.00}{0.01}{0.00}{0.00}{0.00} & 
                              \plotprobs[1.307]{0.54}{0.01}{0.03}{0.27}{0.04}{0.02}{0.05}{0.02} & 
                              \plotprobs[0.923]{0.67}{0.01}{0.23}{0.09}{0.00}{0.00}{0.01}{0.00} & 
                              \plotprobs[0.156]{0.40}{0.02}{0.01}{0.21}{0.22}{0.06}{0.02}{0.06} & 
                              \plotprobs[1.245]{0.54}{0.00}{0.02}{0.25}{0.13}{0.03}{0.02}{0.01} \\                                      
       \trueprobs & \plotprobs[1.735]{0.22}{0.00}{0.11}{0.11}{0.11}{0.22}{0.22}{0.00}
         & \aleprobs			& \plotprobs[0.454]{0.88}{0.09}{0.01}{0.00}{0.02}{0.00}{0.00}{0.00} &
                              \plotprobs[1.068]{0.60}{0.01}{0.02}{0.30}{0.01}{0.02}{0.02}{0.03} &
                              \plotprobs[0.733]{0.76}{0.00}{0.15}{0.07}{0.00}{0.00}{0.00}{0.00} & 
                              \plotprobs[1.469]{0.42}{0.02}{0.03}{0.21}{0.23}{0.06}{0.00}{0.03} & 
                              \plotprobs[0.886]{0.71}{0.00}{0.01}{0.16}{0.11}{0.01}{0.00}{0.00} \\                              
        & & \epiprobs			& &
                              \plotprobs[1.095]{0.62}{0.01}{0.01}{0.26}{0.02}{0.02}{0.02}{0.04} &
                              \plotprobs[0.304]{0.94}{0.00}{0.03}{0.03}{0.00}{0.00}{0.00}{0.00} & 
                              \plotprobs[0.000]{1.00}{0.00}{0.00}{0.00}{0.00}{0.00}{0.00}{0.00} & 
                              \plotprobs[0.493]{0.86}{0.00}{0.00}{0.11}{0.02}{0.00}{0.01}{0.00} \\
    \end{tabular}
    \hspace*{3cm}\begin{tikzpicture}
        \begin{customlegend}[legend columns = 8,legend style = {column sep=1ex}, legend cell align = left,
            legend entries={Neutral, Happiness, Surprise, Sadness, Anger, Disgust, Fear, Contempt}]
            \addlegendimage{mark=none,red, only marks}
            \addlegendimage{mark=none,magenta, only marks}
            \addlegendimage{mark=none,orange, only marks}
            \addlegendimage{mark=none,violet, only marks}
            \addlegendimage{mark=none,green, only marks}
            \addlegendimage{mark=none,lime, only marks}
            \addlegendimage{mark=none,teal, only marks}
            \addlegendimage{mark=none,blue, only marks}
        \end{customlegend}
    \end{tikzpicture}
    \caption{Facial image 2004 from the FER+ test set. Here we show the aleatoric and epistemic probabilities produced by different uncertainty quantification methods. The entropy of each distribution is displayed in the top right corner. There is no dominant correct class and all methods predict Neutral with different levels of confidence.}
    \label{fig:ferplus_probsA}
\end{figure*}

\subsection{Classification in the FERPlus Dataset}

The FERPlus image dataset \cite{ferPlus} is an improved version of the Facial Emotion Recognition (FER) dataset \cite{goodfellow2013challenges}. The task is to classify facial emotions into eight classes, which makes it an inherently ambiguous task. In the FER+ dataset, labels were crowdsourced, and each image received annotations from 10 different annotators, from where a probability distribution can be computed. This scheme partially represents emotional ambiguity in facial images, as seen by the human annotators. We believe this is an excellent example to showcase uncertainty in a classification setting, since the data has aleatoric uncertainty, and the model will have a degree of epistemic uncertainty due to the ambiguity between classes.


We train a simple 8-layer CNN, which has been detailed in the supplementary, using a cross-entropy loss using probability distribution labels (not one-hot encoded) which represent a distribution over eight emotion classes (Neutral, Happiness, Surprise, Sadness, Anger, Disgust, Fear, Contempt). We implement the sampling softmax function as described in Sec \ref{refSectionClasification}, first by using two fully connected layers with $C = 8$ neurons, each predicting a mean and variance for logits, which are given as input to the sampling softmax, which produces the final output probabilities. This is shown in Figure \ref{fig:sampling_softmax_graph}. The mean logit layer uses a linear activation, while the logit variance logit layer uses a softplus activation to produce positive variances. The probabilities are supervised by the cross-entropy loss directly, and automatic differentiation is used to compute gradients through the sampling function.

\begin{figure*}[!ht]
    \footnotesize
    \centering
    \begin{tabular}{llllllll}
        & &      				& Baseline & Dropout & DropConnect & Flipout & Ensembles\\
        \inpimage & \includegraphics[width=0.06\linewidth]{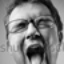}
        & \predprobs			& \plotprobs[1.208]{0.00}{0.04}{0.41}{0.00}{0.31}{0.00}{0.24}{0.00} & 
                                  \plotprobs[1.351]{0.01}{0.01}{0.24}{0.08}{0.49}{0.01}{0.16}{0.01} & 
                                  \plotprobs[1.187]{0.00}{0.07}{0.26}{0.00}{0.50}{0.00}{0.17}{0.00} & 
                                  \plotprobs[1.424]{0.08}{0.03}{0.15}{0.06}{0.55}{0.01}{0.11}{0.01} & 
                                  \plotprobs[1.477]{0.00}{0.04}{0.26}{0.05}{0.37}{0.03}{0.24}{0.00} \\
                    
        \trueprobs & \plotprobs[1.735]{0.00}{0.22}{0.11}{0.22}{0.11}{0.22}{0.11}{0.00}
        & \aleprobs			& \plotprobs[1.241]{0.00}{0.06}{0.37}{0.00}{0.33}{0.00}{0.25}{0.00} &
                              \plotprobs[0.911]{0.00}{0.00}{0.19}{0.01}{0.68}{0.00}{0.12}{0.00} &
                              \plotprobs[1.211]{0.00}{0.14}{0.22}{0.00}{0.51}{0.00}{0.13}{0.00} & 
                              \plotprobs[1.243]{0.03}{0.04}{0.14}{0.03}{0.62}{0.02}{0.12}{0.00} & 
                              \plotprobs[0.996]{0.00}{0.01}{0.19}{0.01}{0.62}{0.00}{0.17}{0.00} \\
        
        & & \epiprobs			& &
                              \plotprobs[0.702]{0.00}{0.00}{0.11}{0.01}{0.80}{0.00}{0.07}{0.00} &
                              \plotprobs[0.145]{0.00}{0.00}{0.01}{0.00}{0.97}{0.00}{0.02}{0.00} & 
                              \plotprobs[0.000]{0.00}{0.00}{0.00}{0.00}{1.00}{0.00}{0.00}{0.00} & 
                              \plotprobs[0.498]{0.00}{0.02}{0.05}{0.01}{0.89}{0.01}{0.02}{0.00} \\
    \end{tabular}
    \hspace*{3cm}\begin{tikzpicture}
        \begin{customlegend}[legend columns = 8,legend style = {column sep=1ex}, legend cell align = left,
            legend entries={Neutral, Happiness, Surprise, Sadness, Anger, Disgust, Fear, Contempt}]
            \addlegendimage{mark=none,red, only marks}
            \addlegendimage{mark=none,magenta, only marks}
            \addlegendimage{mark=none,orange, only marks}
            \addlegendimage{mark=none,violet, only marks}
            \addlegendimage{mark=none,green, only marks}
            \addlegendimage{mark=none,lime, only marks}
            \addlegendimage{mark=none,teal, only marks}
            \addlegendimage{mark=none,blue, only marks}
        \end{customlegend}
    \end{tikzpicture}
    \caption{Facial image 492 from the FER+ test set. Here we show the aleatoric and epistemic probabilities produced by different uncertainty quantification methods. The entropy of each distribution is displayed in the top right corner. Note that there is no dominant correct class, and all methods agree to predict Anger with varying levels of epistemic uncertainty.}
    \label{fig:ferplus_probsB}
\end{figure*}

\begin{figure*}[!ht]
    \footnotesize
    \centering
    \begin{tabular}{llllllll}
    & &      				& Baseline & Dropout & DropConnect & Flipout & Ensembles\\
    \inpimage & \includegraphics[width=0.06\linewidth]{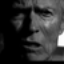}
    & \predprobs			& \plotprobs[0.816]{0.75}{0.00}{0.00}{0.00}{0.00}{0.11}{0.04}{0.10} & 
                              \plotprobs[1.123]{0.65}{0.00}{0.01}{0.11}{0.14}{0.07}{0.00}{0.02} & 
                              \plotprobs[0.815]{0.70}{0.00}{0.00}{0.24}{0.04}{0.02}{0.00}{0.00} & 
                              \plotprobs[1.793]{0.24}{0.03}{0.14}{0.09}{0.33}{0.06}{0.06}{0.05} & 
                              \plotprobs[1.088]{0.63}{0.00}{0.00}{0.22}{0.08}{0.05}{0.01}{0.02} \\                                      
                              
      \trueprobs 		& \plotprobs[1.696]{0.30}{0.00}{0.00}{0.10}{0.20}{0.10}{0.20}{0.00}
    & \aleprobs			& \plotprobs[0.799]{0.75}{0.00}{0.00}{0.00}{0.00}{0.15}{0.06}{0.04} &
                          \plotprobs[1.116]{0.65}{0.00}{0.01}{0.11}{0.14}{0.08}{0.00}{0.02} &
                          \plotprobs[0.583]{0.74}{0.00}{0.00}{0.26}{0.00}{0.00}{0.00}{0.00} & 
                          \plotprobs[1.635]{0.31}{0.01}{0.13}{0.05}{0.35}{0.04}{0.07}{0.04} & 
                          \plotprobs[0.823]{0.72}{0.00}{0.00}{0.16}{0.10}{0.01}{0.00}{0.00} \\                              
                          
    & & \epiprobs			& &
                          \plotprobs[0.712]{0.80}{0.00}{0.00}{0.04}{0.13}{0.02}{0.00}{0.01} &
                          \plotprobs[0.035]{0.99}{0.00}{0.00}{0.01}{0.00}{0.00}{0.00}{0.00} & 
                          \plotprobs[0.000]{0.00}{0.00}{0.00}{0.00}{1.00}{0.00}{0.00}{0.00} & 
                          \plotprobs[0.331]{0.92}{0.00}{0.00}{0.01}{0.06}{0.00}{0.00}{0.00} \\
    \end{tabular}        
    \hspace*{3cm}\begin{tikzpicture}
        \begin{customlegend}[legend columns = 8,legend style = {column sep=1ex}, legend cell align = left,
            legend entries={Neutral, Happiness, Surprise, Sadness, Anger, Disgust, Fear, Contempt}]
            \addlegendimage{mark=none,red, only marks}
            \addlegendimage{mark=none,magenta, only marks}
            \addlegendimage{mark=none,orange, only marks}
            \addlegendimage{mark=none,violet, only marks}
            \addlegendimage{mark=none,green, only marks}
            \addlegendimage{mark=none,lime, only marks}
            \addlegendimage{mark=none,teal, only marks}
            \addlegendimage{mark=none,blue, only marks}
        \end{customlegend}
    \end{tikzpicture}
    \caption{Facial image 2791 from the FER+ test set. Here we show the aleatoric and epistemic probabilities produced by different uncertainty quantification methods. The entropy of each distribution is displayed in the top right corner. All methods except Flipout predict the correct class (Neutral) with varying levels of uncertainty. Flipout predicts zero epistemic uncertainty for an incorrect prediction.}
    \label{fig:ferplus_probsC}
\end{figure*}

\begin{figure*}[t]
    \footnotesize
    \centering
    \begin{tabular}{llllllll}
        & &      				& Baseline & Dropout & DropConnect & Flipout & Ensembles\\
        \inpimage & \includegraphics[width=0.06\linewidth]{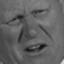}
        & \predprobs			& \plotprobs[1.414]{0.00}{0.00}{0.00}{0.15}{0.36}{0.28}{0.19}{0.02} & 
        \plotprobs[1.421]{0.09}{0.02}{0.04}{0.33}{0.44}{0.05}{0.02}{0.01} & 
        \plotprobs[1.186]{0.00}{0.00}{0.20}{0.06}{0.53}{0.20}{0.00}{0.00} & 
        \plotprobs[0.984]{0.07}{0.74}{0.07}{0.07}{0.01}{0.00}{0.02}{0.02} & 
        \plotprobs[1.326]{0.09}{0.03}{0.01}{0.50}{0.28}{0.06}{0.02}{0.00} \\                                      
        
        \trueprobs 		& \plotprobs[1.667]{0.38}{0.12}{0.12}{0.12}{0.12}{0.12}{0.00}{0.00}
        & \aleprobs			& \plotprobs[1.399]{0.00}{0.00}{0.02}{0.12}{0.35}{0.17}{0.34}{0.01} &
        \plotprobs[1.496]{0.12}{0.02}{0.04}{0.30}{0.42}{0.08}{0.02}{0.01} &
        \plotprobs[1.047]{0.00}{0.00}{0.16}{0.04}{0.60}{0.20}{0.00}{0.00} & 
        \plotprobs[1.179]{0.04}{0.69}{0.06}{0.09}{0.04}{0.01}{0.04}{0.03} & 
        \plotprobs[0.944]{0.07}{0.00}{0.00}{0.66}{0.22}{0.05}{0.00}{0.00} \\                              
        
        & & \epiprobs			& &
        \plotprobs[1.588]{0.11}{0.03}{0.06}{0.28}{0.40}{0.08}{0.02}{0.01} &
        \plotprobs[0.251]{0.00}{0.00}{0.00}{0.02}{0.94}{0.04}{0.00}{0.00} & 
        \plotprobs[0.000]{0.00}{1.00}{0.00}{0.00}{0.00}{0.00}{0.00}{0.00} & 
        \plotprobs[0.972]{0.09}{0.02}{0.00}{0.73}{0.10}{0.04}{0.02}{0.00} \\
    \end{tabular}
    \hspace*{3cm}\begin{tikzpicture}
        \begin{customlegend}[legend columns = 8,legend style = {column sep=1ex}, legend cell align = left,
            legend entries={Neutral, Happiness, Surprise, Sadness, Anger, Disgust, Fear, Contempt}]
            \addlegendimage{mark=none,red, only marks}
            \addlegendimage{mark=none,magenta, only marks}
            \addlegendimage{mark=none,orange, only marks}
            \addlegendimage{mark=none,violet, only marks}
            \addlegendimage{mark=none,green, only marks}
            \addlegendimage{mark=none,lime, only marks}
            \addlegendimage{mark=none,teal, only marks}
            \addlegendimage{mark=none,blue, only marks}
        \end{customlegend}
    \end{tikzpicture}
    \caption{Facial image 3100 from the FER+ test set. Here we show the aleatoric and epistemic probabilities produced by different uncertainty quantification methods. The entropy of each distribution is displayed in the top right corner. No method predicts the correct class (neutral), with Flipout having zero epistemic uncertainty and Ensembles/Dropout having the highest epistemic one.}
    \label{fig:ferplus_probsD}
\end{figure*}

\begin{figure*}[t]
    \footnotesize
    \centering
    \begin{tabular}{llllllll}
        & &      				& Baseline & Dropout & DropConnect & Flipout & Ensembles\\
        \inpimage & \includegraphics[width=0.06\linewidth]{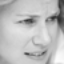}
        & \predprobs			& \plotprobs[1.558]{0.15}{0.03}{0.25}{0.34}{0.18}{0.05}{0.00}{0.00} & 
        \plotprobs[1.629]{0.27}{0.27}{0.05}{0.29}{0.05}{0.03}{0.02}{0.02} & 
        \plotprobs[1.556]{0.06}{0.07}{0.15}{0.31}{0.35}{0.05}{0.00}{0.00} & 
        \plotprobs[1.515]{0.14}{0.09}{0.06}{0.09}{0.52}{0.04}{0.00}{0.06} & 
        \plotprobs[1.648]{0.28}{0.23}{0.06}{0.27}{0.11}{0.05}{0.01}{0.00} \\                                      
        
        \trueprobs 		& \plotprobs[1.643]{0.30}{0.10}{0.10}{0.30}{0.10}{0.10}{0.00}{0.00}
        & \aleprobs			& \plotprobs[1.519]{0.18}{0.02}{0.27}{0.28}{0.24}{0.02}{0.00}{0.00} &
        \plotprobs[1.402]{0.27}{0.25}{0.02}{0.40}{0.03}{0.01}{0.01}{0.02} &
        \plotprobs[1.456]{0.04}{0.08}{0.16}{0.25}{0.44}{0.04}{0.00}{0.00} & 
        \plotprobs[1.395]{0.24}{0.04}{0.06}{0.09}{0.52}{0.01}{0.01}{0.03} & 
        \plotprobs[1.393]{0.28}{0.26}{0.03}{0.37}{0.05}{0.01}{0.00}{0.00} \\                              
        
        & & \epiprobs			& &
        \plotprobs[1.462]{0.23}{0.26}{0.02}{0.40}{0.03}{0.03}{0.01}{0.02} &
        \plotprobs[0.508]{0.00}{0.00}{0.00}{0.81}{0.18}{0.00}{0.00}{0.00} & 
        \plotprobs[0.000]{0.00}{0.00}{0.00}{0.00}{1.00}{0.00}{0.00}{0.00} & 
        \plotprobs[1.516]{0.40}{0.14}{0.02}{0.30}{0.06}{0.04}{0.04}{0.00} \\
    \end{tabular}
    \hspace*{3cm}\begin{tikzpicture}
        \begin{customlegend}[legend columns = 8,legend style = {column sep=1ex}, legend cell align = left,
            legend entries={Neutral, Happiness, Surprise, Sadness, Anger, Disgust, Fear, Contempt}]
            \addlegendimage{mark=none,red, only marks}
            \addlegendimage{mark=none,magenta, only marks}
            \addlegendimage{mark=none,orange, only marks}
            \addlegendimage{mark=none,violet, only marks}
            \addlegendimage{mark=none,green, only marks}
            \addlegendimage{mark=none,lime, only marks}
            \addlegendimage{mark=none,teal, only marks}
            \addlegendimage{mark=none,blue, only marks}
        \end{customlegend}
    \end{tikzpicture}
    \caption{Facial image 813 from the FER+ test set. Here we show the aleatoric and epistemic probabilities produced by different uncertainty quantification methods. The entropy of each distribution is displayed in the top right corner. Ensembles has the highest epistemic uncertainty, while Flipout has zero and predicts an incorrect class.}
    \label{fig:ferplus_probsE}
\end{figure*}

We report test loss and accuracy in Table \ref{fig:ferplus_metrics}. For a qualitative evaluation, we select the top 5 images on the test set according to the entropy computed from the ground truth class probability distribution. Then we compute the disentangled aleatoric and epistemic logits using the method described in Sec \ref{refSectionClasification}. Finally, we transform them into probabilities using the mean and variance through the sampling softmax function. We present these results in Figures \ref{fig:ferplus_probsA} to \ref{fig:ferplus_probsE}. Each figure shows the entropy of each probability distribution, as a measure of uncertainty. It should be noted that, unlike regression, the aleatoric and epistemic probabilities and entropy values do not add to the total predicted entropy.

Flipout seems to produce zero epistemic uncertainty for all the selected examples, which is odd but consistent with the toy regression example. We believe this is due to Flipout's powerful variance reduction effect, which seems to affect epistemic uncertainty in particular, putting all uncertainty mass into aleatoric uncertainty.

In most cases, the aleatoric probabilities follow or are similar to the ground truth probabilities. There is a more significant disagreement between epistemic uncertainty across different uncertainty quantification methods, which is unexpected, as all methods use the same network topology, except for some differences in how specific UQ methods are applied to the architecture (\eg replacing layers, adding Dropout layers). We believe these results show that adding UQ to a neural network has a more considerable impact on its epistemic uncertainty than the aleatoric one.

Table \ref{fig:ferplus_metrics} presents accuracy and cross-entropy metrics in the FER+ dataset. The Ensembles approach is the evident best in terms of accuracy, but Dropout has the smallest loss, closely followed by ensembles. Accuracy is only evaluated for the highest probability prediction, and the ground truth probabilities in FER+ for many cases do not have a clear dominating class (for example in Figures \ref{fig:ferplus_probsA}, \ref{fig:ferplus_probsB}), while the cross-entropy loss measures a fit in terms of probability distributions.

\subsection{Discussion}

Our expectation is that aleatoric uncertainty should be independent of the uncertainty quantification method, while epistemic uncertainty should depend on the specific uncertainty method. This is because aleatoric uncertainty is associated to the data, so the model performs regression of the variance in the data, while epistemic is associated to the model itself, and the uncertainty quantification method interacts with the model directly. Both kinds of uncertainty should not interact with each other.

Our classification and regression results mainly show that the uncertainty quantification method does affect both aleatoric and epistemic uncertainty quantification, with some surprising results. As mentioned before, this is expected for epistemic but not for aleatoric uncertainty.

In particular, Flipout seems to confuse aleatoric and epistemic uncertainty, producing almost zero epistemic uncertainty in both classification and regression settings. This behavior is unexpected and most likely incorrect, and we would not recommend its use if disentangling uncertainty is an application requirement.

Some uncertainty models like Dropout and DropConnect are heavily affected by the use of the $\beta$-NLL loss in a regression setting, which improves aleatoric uncertainty estimation, but seems to also affect epistemic uncertainty, mostly in a positive way. However, not all methods equally benefit. It is the case of Flipout, which seems to have smaller epistemic uncertainty, with little change to aleatoric uncertainty.

Overall the best combination of uncertainty method and loss seems to be Ensembles with the $\beta$-NLL loss for regression, and ensembles for classification, obtaining the best accuracy in our FER+ dataset experiments, and a good disentangling of aleatoric and epistemic uncertainty in our toy regression experiments.

We believe that these insights will help the community and practitioners select appropriate uncertainty quantification methods, and expand the use of disentanglement of aleatoric and epistemic uncertainty.

\section{Conclusions and Future Work}

In this paper, we generalized methods to disentangle aleatoric and epistemic uncertainty, and compared different uncertainty quantification methods in terms of their disentangling quality. 
Our results show the differences between using different uncertainty quantification methods. We illustrate how aleatoric and epistemic uncertainties interact with each other, which is unexpected and partially violates the definitions of each kind of uncertainty, specially for aleatoric uncertainty. In particular, Flipout seems to produce almost zero epistemic uncertainty, putting all mass into aleatoric uncertainty. Ensembles unsurprisingly seem to have the best disentangling quality, when trained using the $\beta$-NLL loss for regression, and with the sampling softmax function for classification.

We also explored the approximation error produced by the number of samples used in the sampling softmax function, and we show that at least $N = 100$ samples are required for a good approximation and to have a misclassification error close to zero.

We expect that our results can guide practitioners to select uncertainty quantification methods, include disentangling quality as an additional metric for future uncertainty quantification methods, and expand the use of disentangled uncertainty quantification methods.

We believe that our results show the need for additional research to understand how epistemic and aleatoric uncertainty interact when estimated using machine learning models, and ways to reduce interactions for aleatoric uncertainty.

\clearpage
\FloatBarrier
{\small
\bibliographystyle{ieee_fullname}
\bibliography{egbib}
}

\appendix
\clearpage
\onecolumn
\section{Training Data for Toy Regression Example}

\begin{figure}[!h]
    \centering
    \includegraphics[width=0.49\linewidth]{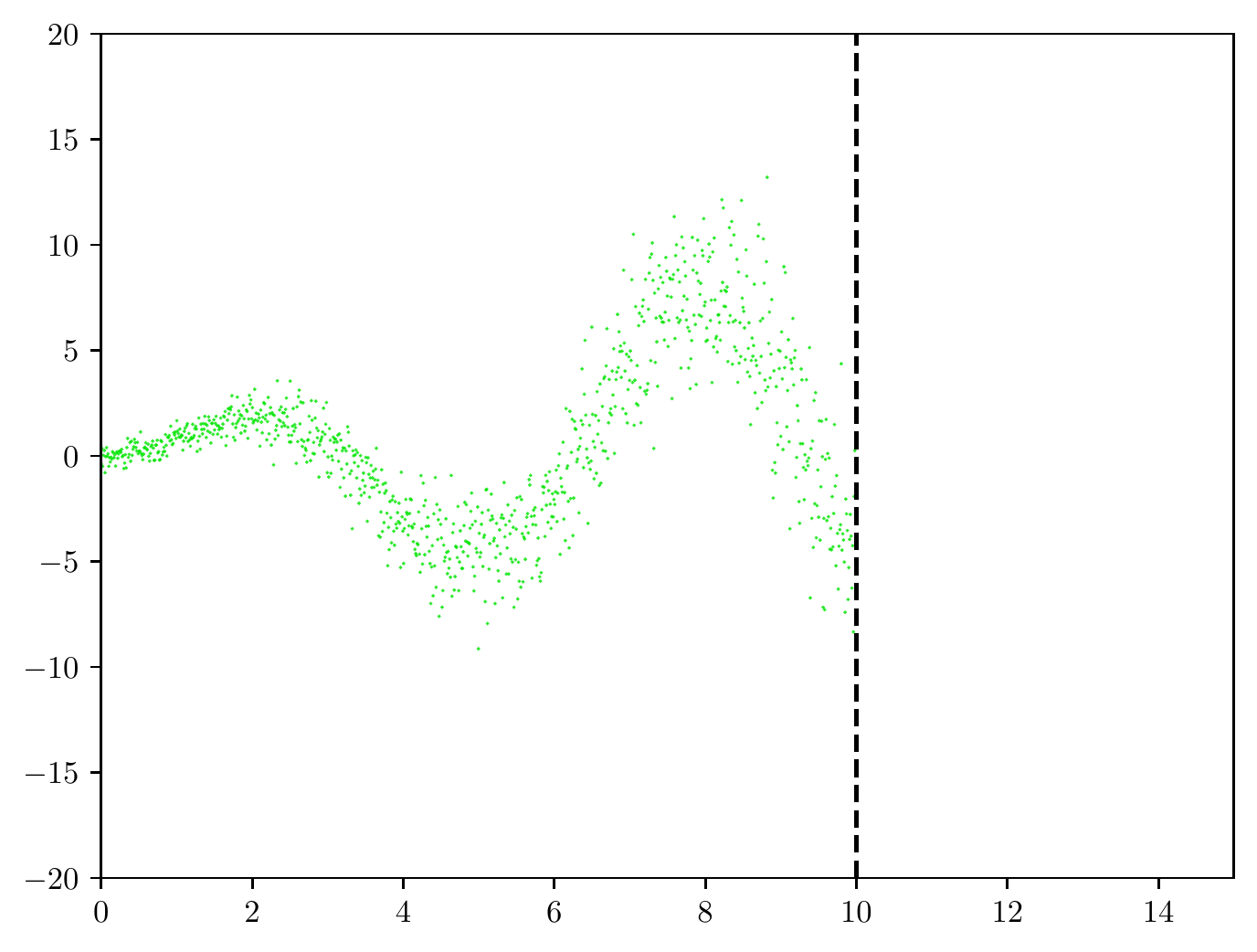}
    \caption{Training data used for the toy sinusoid problem with heteroscedatic aleatoric uncertainty.}
    \label{fig:toy_regression_data}
\end{figure}

In this section we present the training samples used for evaluation of the toy regression example, presented in Figure \ref{fig:toy_regression_data}. This data clearly shows how aleatoric uncertainty varies with the input (it increases linearly with $x$), which is denominated heteroscedatic uncertainty.

\FloatBarrier

\section{Neural Network Architecture Details}

\begin{enumerate}
    \item{\textbf{Regression task (Sinusoidal function with noise):}
        
    - General configurations for all networks: Number of epochs: 700. Batch size: 32. Optimizer: Adam ($lr=0.001$, $\beta_1=0.9$, and $\beta_2=0.999$).
    
    - Configuration applied at each of the methods tested:
    
    \begin{itemize}
        \item{\textbf{Baseline}. Dense(32, ReLU) - Dense(32, ReLU).}
        \item{\textbf{Dropout}. Dense(32, ReLU) - Dropout(0.25) - Dense(32, ReLU) - Dropout(0.25).}
        \item{\textbf{DropConnect}. DropConnectDense(32, ReLU, $p = 0.1$) - DropConnectDense(32, ReLU, $p = 0.1$).}
        \item{\textbf{Flipout}. FlipoutDense(32, ReLU) - FlipoutDense(32, ReLU). Prior is disabled.}
        \item{\textbf{Ensembles}. 5 copies of the Baseline model trained with different random weight initializations.}
    \end{itemize}

        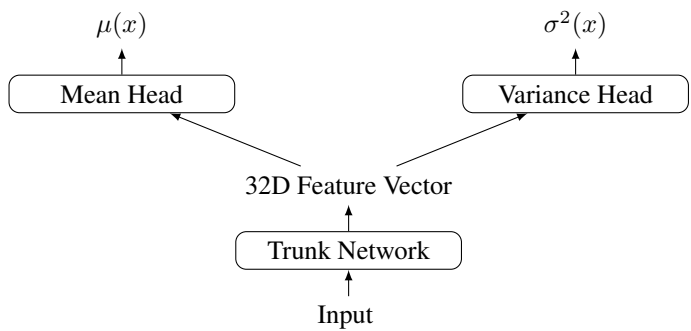
\begin{figure}[h]
        \centering
        \begin{tikzpicture}[style={align=center, minimum height=0.5cm, minimum width = 3.0cm}]
            \node[] (dummy) {};
            \node[rounded corners, draw, left=0.001em of dummy] (objClassifier) {{Mean Head}};
            \node[above=1em of objClassifier] (objClass) {{$\mu(x)$}};
            
            \node[rounded corners, draw, right=0.01em of dummy] (objDetector) {{Variance Head}};
            \node[above=1em of objDetector] (obj) {{$\sigma^2(x)$}};
            
            \node[below=2em of dummy] (fc) {{32D Feature Vector}};
            
            \node[rounded corners, draw, below=1em of fc](conv) {{Trunk Network}};
            \node[below=1em of conv](inputImage) {{Input }};
            \draw[-latex] (inputImage) -- (conv);
            \draw[-latex] (conv) -- (fc);
            \draw[-latex] (fc) -- (objDetector);
            \draw[-latex] (fc) -- (objClassifier);
            \draw[-latex] (objClassifier) -- (objClass);
            \draw[-latex] (objDetector) -- (obj);
        \end{tikzpicture}
        \caption{Diagram of the basic network architecture for regression with uncertainty. Each trunk network is a specific implementation of an uncertainty quantification method, as shown in the list above.}
        \label{fig:reg_unc_twoheads}
    \end{figure}

    - Each of the networks displayed above are trunk networks. To predict regression values with uncertainty, two parallel layers are added. One Dense(1, Linear) for predicting the mean $\mu(x)$, and one Dense(1, Softplus) to predict the standard deviation $\sigma(x)$. This network configuration is visually displayed in Figure \ref{fig:reg_unc_twoheads}.
    }

    \item{\textbf{Classification task (FER+ Dataset):}
    
    - General configurations for all networks: Number of epochs: 120. Batch size: 64. Optimizer: Adam ($lr=0.001$, $\beta_1=0.9$, and $\beta_2=0.999$).
    
    - Configuration applied at each of the methods tested:
    
    \begin{itemize}
        \item{\textbf{Baseline}. Conv2D(64, 3 $\times$ 3, ReLU) - BatchNorm() - MaxPool2D(2 $\times$ 2) - Conv2D(128, 3 $\times$ 3, ReLU) - BatchNorm() - MaxPool2D(2 $\times$ 2) - Conv2D(128, 3 $\times$ 3, ReLU) - BatchNorm() - MaxPool2D(2 $\times$ 2) - Flatten() - Dense(256, ReLU) - Dense(256, ReLU).}
        \item{\textbf{Dropout}. Conv2D(64, 3 $\times$ 3, ReLU) - BatchNorm() - MaxPool2D(2 $\times$ 2) - Conv2D(128, 3 $\times$ 3, ReLU) - BatchNorm() - MaxPool2D(2 $\times$ 2) - Conv2D(128, 3 $\times$ 3, ReLU) - BatchNorm() - MaxPool2D(2 $\times$ 2) - Flatten() - StochasticDropout($p=0.25$) - Dense(256, ReLU) - StochasticDropout($p=0.25$) - Dense(256, ReLU).}
        \item{\textbf{DropConnect}. Conv2D(64, 3 $\times$ 3, ReLU) - BatchNorm() - MaxPool2D(2 $\times$ 2) - Conv2D(128, 3 $\times$ 3, ReLU) - BatchNorm() - MaxPool2D(2 $\times$ 2) - Conv2D(128, 3 $\times$ 3, ReLU) - BatchNorm() - MaxPool2D(2 $\times$ 2) - Flatten() - DropConnectDense(256, ReLU, $p=0.10$) - DropConnectDense(256, ReLU, $p=0.10$).}
        \item{\textbf{Flipout}. Conv2D(64, 3 $\times$ 3, ReLU) - BatchNorm() - MaxPool2D(2 $\times$ 2) - Conv2D(128, 3 $\times$ 3, ReLU) - BatchNorm() - MaxPool2D(2 $\times$ 2) - Conv2D(128, 3 $\times$ 3, ReLU) - BatchNorm() - MaxPool2D(2 $\times$ 2) - Flatten() - FlipoutDense(256, ReLU, $p=0.10$) - FlipoutDense(256, ReLU, $p=0.10$).}
        \item{\textbf{Ensembles}. 5 copies of the Baseline model trained with different random weight initializations.}
    \end{itemize}           
    }
\end{enumerate}

\end{document}